\documentclass[letterpaper]{article} 
\usepackage{aaai2027}  
\nocopyright
\usepackage[hyphens]{url}  
\usepackage{graphicx} 
\usepackage{natbib}  
\usepackage{caption} 
\usepackage{algorithm}
\usepackage{algorithmic}

\usepackage{booktabs}
\usepackage{multirow}
\usepackage{graphicx}
\usepackage{tabularx}
\usepackage{array}
\usepackage{newfloat}
\usepackage{amsfonts}
\usepackage{amsmath}
\usepackage{listings}
\DeclareCaptionStyle{ruled}{labelfont=normalfont,labelsep=colon,strut=off} 
\floatstyle{ruled}
\newfloat{listing}{tb}{lst}{}
\floatname{listing}{Listing}

\usepackage{booktabs}

\title{Revisiting the Current Frame: \\Physical-Trace-Guided Network Output Correction for Video Restoration}
\author {
    Yifeng Lin\textsuperscript{\rm 1},
    Liuxiang Qiu\textsuperscript{\rm 1},
    Guangming Ren\textsuperscript{\rm 1},
    Tiesong Zhao\textsuperscript{\rm 1}\corresponding
}
\affiliations {
    \textsuperscript{\rm 1}Fujian Key Lab for Intelligent Processing and Wireless Transmission of Media Information, College of Physics and Information Engineering, Fuzhou University\\
    241120007@fzu.edu.cn, liuxiangqiu007@gmail.com, 241120121@fzu.edu.cn, t.zhao@fzu.edu.cn
}

\begin{document}

\maketitle

\begin{abstract}
Video restoration methods exploit temporal information to recover information missing from degraded observations. However, reference frames within the sequence may introduce inconsistent degradation, content discrepancy, or reconstruction errors due to physical image-formation variations, occlusion, and imperfect temporal aggregation. Existing approaches mainly focus on improving restoration networks, while the reliability of the generated outputs at different spatial locations remains largely unexplored. In this work, we propose ANCHOR, a model-agnostic framework that revisits the low-quality current frame as a temporally aligned anchor for video restoration correction. Specifically‌, ANCHOR estimates a spatial trust field from heterogeneous physical-trace evidence and adaptively balances the restoration proposal with the original observation. Experiments on High Dynamic Range video reconstruction and video deraining demonstrate consistent improvements across various state-of-the-art restoration models, validating the effectiveness of reliability-aware output correction for video restoration.
\end{abstract}


\section{Introduction}

\begin{figure}[!ht]
  \begin{center}
    \centerline{\includegraphics[width=\columnwidth]{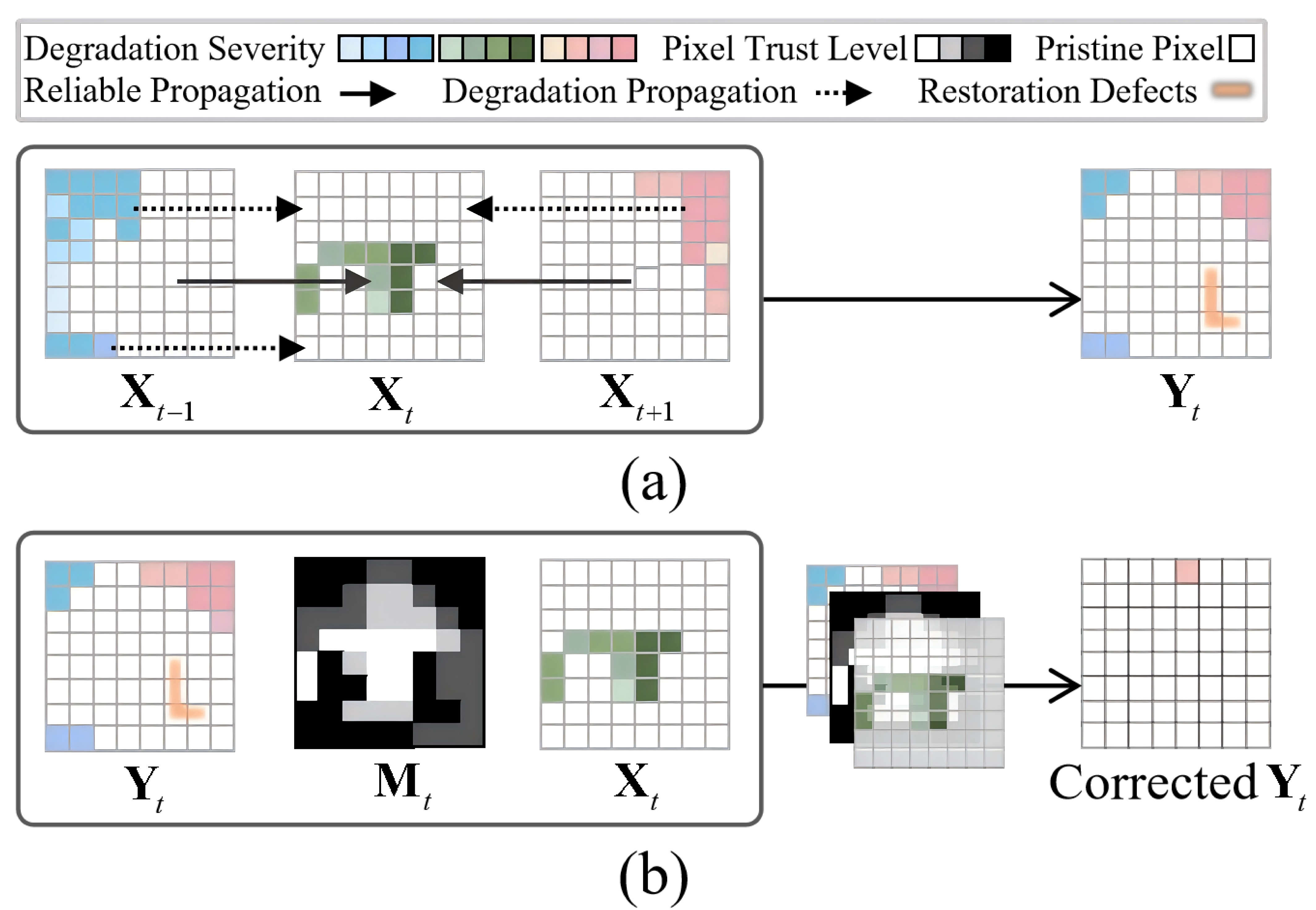}}
\caption{(a) Aggregating reference frames $\mathbf{X}_{t\pm1}$ with the current frame $\mathbf{X}_t$ may introduce locally unreliable modifications. (b) ANCHOR predicts a correction field $\mathbf{M}_t$ to refine the restoration proposal $\mathbf{Y}_t$ toward the current-frame anchor.}
    \label{caotu}
  \end{center}
\end{figure}

Video restoration exploits temporal information to recover missing or corrupted content in the target frame. Existing methods aggregate reference-frame information through temporal propagation \cite{tassano2019dvdnet}, attention \cite{liang2022recurrent,liang2024vrt}, or state-space modeling \cite{guo2024mambair,guo2025mambairv2}. However, reference frames can be unreliable due to content variation, occlusion, inconsistent degradation, and alignment errors. This issue is particularly pronounced for physical image-formation degradations, including exposure integration \cite{cui2024exposure}, dynamic occlusion \cite{sun2026delivr}, and reflection \cite{he2025rethinking}, where temporal observations may provide complementary information while introducing inconsistent evidence. Consequently, temporal aggregation may produce locally unreliable modifications in the restored output, as illustrated in Fig.~\ref{caotu}.

Existing methods mitigate restoration-induced errors through robust alignment and selective fusion \cite{li2024neural,Yue_2026_CVPR}, temporal consistency regularization \cite{zhou2024upscale,zhu2024temporally}, photometric or color calibration \cite{ye2024deep,wang2025retinexmcnet}, and dedicated post-processing approaches that refine restored outputs using learned priors or artifact-specific models \cite{lei2023blind,ali2023task}. However, most of these approaches are task-specific, integrated into the restoration pipeline, or designed for specific artifacts such as ghosting, flickering, and color shifts. Few methods explicitly revisit the current low-quality frame as a spatially aligned observation reference to correct the output of an arbitrary pretrained restoration network. Unlike reference frames, the current frame is inherently aligned with the target instant and may preserve more faithful radiometric and structural information in mildly degraded regions. Therefore, we treat it not only as an input, but also as a current-frame observation anchor for output correction. This motivates a fundamental question:

\textit{Where should the temporal reconstruction be trusted and corrected using the current-frame observation?}

To address this question, we propose ANCHOR, an Adaptive Network-output Correction framework based on Heterogeneous Observation Reliability. ANCHOR treats the restoration output as a temporal proposal and the current frame as an observation anchor. It extracts reliability evidence from radiometric observability, structural preservation, and temporal consistency, and integrates these cues through evidence-preserving dense consensus. The resulting spatial trust field enables reliability-aware correction between the proposal and anchor, yielding adaptive restoration refinement. In summary, our contributions are as follows:

\begin{itemize}
\item We revisit the current frame as an observation anchor for video restoration and formulate a reliability-aware output correction paradigm based on spatial trust estimation.

\item We propose ANCHOR, a physical-trace-guided correction framework that extracts heterogeneous reliability evidence and performs evidence-preserving consensus reasoning for trust estimation.

\item We validate ANCHOR on five datasets for  HDR video reconstruction and video deraining under distinct physical image-formation degradations.
\end{itemize}

\section{Related work}

\subsection{Physics-Induced Video Restoration}
Physical image-formation degradations arise before digitization, when scene radiance is altered by exposure integration, propagation media, dynamic particles, or additional optical paths. Representative tasks include HDR video reconstruction~\cite{yan2019attention,Yue_2026_CVPR}, video deraining~\cite{sun2025semi,sun2026delivr}, video desnowing~\cite{chen2023snow,chen2026density}, video dehazing~\cite{fan2025depth,deng2026towards}, and video reflection removal~\cite{hong2024light,he2025rethinking}. Unlike digital degradations with predefined operators, these degradations are often spatially and temporally varying, content dependent, and coupled with scene motion, resulting in heterogeneous information loss and interference across frames. Existing methods therefore incorporate task-specific physical priors and temporal aggregation mechanisms to exploit complementary cross-frame observations for restoration~\cite{wu2024semi,fan2024driving,zhang2025egvd}. In this work, we conduct experiments on two representative tasks: HDR video reconstruction and video deraining.

\subsection{Restoration Refinement}

Existing refinement methods follow several paradigms. Multistage networks progressively refine predictions within task-specific architectures~\cite{zamir2021multi}, while plug-and-play methods incorporate pretrained priors into iterative solvers constrained by explicit data-fidelity models~\cite{zhang2021plug,zhu2023denoising}. Learned refinement modules provide another direction: DiffGAR suppresses generative artifacts using simulated degradations~\cite{yin2022diffgar}, whereas Ryou and co-authors train an output refiner with perceptually enhanced supervision for real-world image restoration~\cite{ryou2025beyond}. Despite sharing the broad goal of restoration refinement, these methods are developed under distinct problem formulations and, in practice, typically depend on task-specific architectures, forward models, artifact synthesis schemes, or dedicated supervision pipelines. In contrast, ANCHOR provides low-overhead, observation-conditioned refinement without altering existing video restorers and directly reuses their original supervision pipelines. It uses the synchronized current frame to selectively correct output modifications unsupported by aligned observations.

\section{Method}

\begin{figure*}[!ht]
  \centering
  \includegraphics[width=\textwidth]{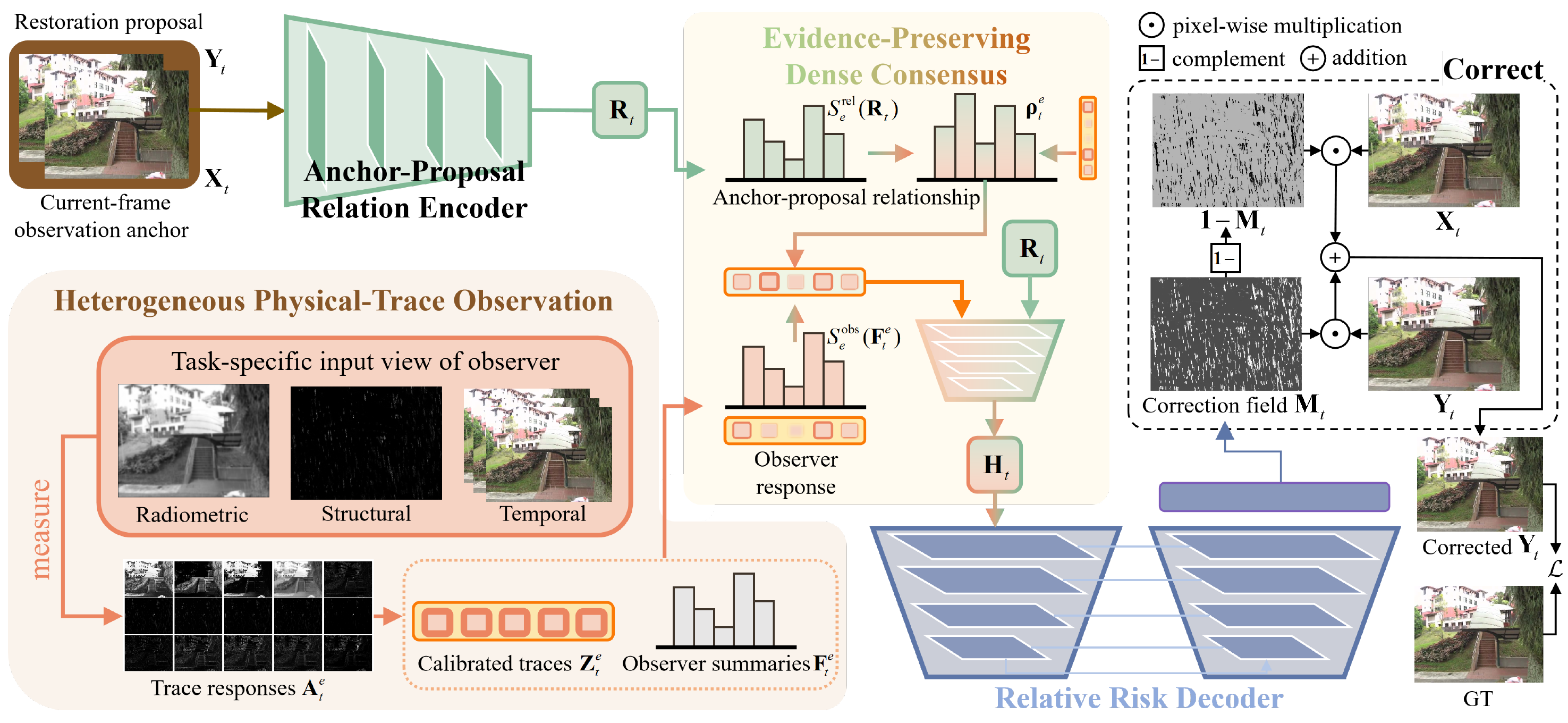}
 \caption{Overview of ANCHOR-Correct. The anchor--proposal relation is encoded into relation-aware features ${\bf R}_t$, which guide the consensus of heterogeneous physical traces to form the enhanced representation ${\bf H}_t$. The resulting representation is decoded into a spatial correction field ${\bf M}_t$ for selectively combining the restoration proposal ${\bf Y}_t$ with the current-frame anchor ${\bf X}_t$.}
  \label{fig:framework}
\end{figure*}

\subsection{Motivation}

Given ${{\cal X}_t}
=
\{{{\bf{X}}_{t-K}},\ldots,{{\bf{X}}_t},\ldots,{{\bf{X}}_{t+K}}\}$, a low-quality video clip centered at time step $t$ with temporal half-window size $K$, where ${{\bf{X}}_t}$ denotes the low-quality frame at time step $t$. Let $F_\phi$ denote a video restoration network parameterized by $\phi$, which aggregates temporal observations and produces a restoration proposal:
\begin{equation}
{{\bf{Y}}_t}=F_\phi({{\cal X}_t}).
\end{equation}
Here, ${{\bf{Y}}_t}$ represents the temporal restoration proposal, and the objective of ANCHOR is to recover the ideal frame ${{\bf{G}}_t}$.

Reference frames provide complementary information to recover missing content, but may also introduce inconsistent modifications due to exposure variations, dynamic occlusions, optical interactions, and other physical effects. Consequently, the restoration proposal may contain unreliable temporal modifications. ANCHOR addresses this issue by introducing the original center frame ${{\bf{X}}_t}$ as a current-frame observation anchor. Although affected by the original degradation, the anchor retains direct observations from the target time step without temporal aggregation, providing complementary evidence for evaluating the reliability of the restoration proposal. Rather than reconstructing the target frame independently, ANCHOR predicts a spatially adaptive correction field ${{\bf{M}}_t}$ to refine the final proposal ${\widehat {\bf{Y}}_t}$ along the anchor--proposal discrepancy direction:
\begin{equation}
{\widehat {\bf{Y}}_t}
=
{{\bf{X}}_t}
+
{{\bf{M}}_t}
\odot
({{\bf{Y}}_t}-{{\bf{X}}_t}),
\label{eq2}
\end{equation}
where $\odot$ represents pixel-wise multiplication. This formulation enables reliability-aware correction: the proposal is adaptively shifted toward the anchor where direct observations are trustworthy, while temporal restoration is preserved where the anchor lacks sufficient evidence.

\subsection{Overall Framework}

The proposed physical-trace guided network output correction framework, named ANCHOR, is illustrated in Fig.~\ref{fig:framework}. ANCHOR addresses a complementary problem of restoration reliability by modeling the relationship between the current-frame observation anchor ${{\bf{X}}_t}$ and the temporal restoration proposal ${{\bf{Y}}_t}$. To model how the restoration network modifies the current observation, we first define the anchor--proposal discrepancy as
${{\bf{\Delta }}_t}
=
{{\bf{Y}}_t}-{{\bf{X}}_t}$. The current-frame anchor, temporal restoration proposal, modification direction, and modification magnitude are concatenated along the channel dimension to get  anchor--proposal relation input ${{\bf{C}}_t}
=
[
{{\bf{X}}_t},
{{\bf{Y}}_t},
{{\bf{\Delta }}_t},
|
{{\bf{\Delta }}_t}
|
]$,where $[\cdot]$ denotes channel-wise concatenation. The four components respectively provide the direct observation at the target time step, the restoration proposal generated by temporal aggregation, the direction of network-induced modification, and its spatial magnitude. 

The Anchor--Proposal Relation Encoder (APRE) transforms ${{\bf{C}}_t}$ into relation-aware features ${{\bf{R}}_t}$, which preserve output-dependent information for subsequent physical reliability reasoning. During relation encoding, Heterogeneous Physical-Trace Observation (HPTO) extract complementary reliability evidence from the low quality video and the anchor--proposal relationship. Building upon them, Evidence-Preserving Dense Consensus (EPDC) estimates spatial routing weights among different trace families and generates the enhanced representation ${\bf{H}}_t$. Subsequently, the Relative-Risk Decoder (RRD) decodes ${\bf{H}}_t$ into a single channel current frame correction field ${{\bf{M}}_t}$. Finally, the predicted correction field ${{\bf{M}}_t}$ is applied along the residual direction between the temporal restoration proposal and the anchor. 

\subsection{Heterogeneous Physical-Trace Observation}

Given heterogeneous trace observers ${\cal E}=\{r,d,\tau\}$, namely the radiometric trace, structural trace, and temporal consistency trace, each observer $e\in{\cal E}$ receives a task-specific input view ${\bf{V}}_t^e$. The radiometric trace leverages the current-frame anchor to evaluate radiometric response and local information preservation. The structural trace exploits the anchor--proposal discrepancy to capture structural changes and restoration-induced modifications. The temporal consistency trace incorporates the current and neighboring frames to measure cross-frame consistency. Each observer applies $J$ physically motivated analytic operators ${{\bf{\Phi }}^e}
=
\{\phi_1^e,\phi_2^e,\ldots,\phi_J^e\}$ to extract interpretable trace responses ${\bf{A}}_t^e$:
\begin{equation}
{\bf{A}}_t^e
=
{{\bf{\Phi }}^e}({\bf{V}}_t^e)
=
\{\phi_j^e({\bf{V}}_t^e)\}_{j=1}^{J}.
\end{equation}

Different observers characterize distinct reliability factors through their measurements. The radiometric trace analyzes extreme intensity responses, local illumination variations, and local contrast to evaluate radiometric preservation. The structural trace captures structural and texture variations using multi-scale high-frequency responses, Laplacian responses, and local variance. The temporal consistency trace measures temporal compatibility through bidirectional frame differences, three-frame median deviation, second-order temporal differences, and temporal variance.

These responses are not physical quantities themselves, but image-domain traces left by physical degradations during imaging. To improve adaptability across different scenes and restoration tasks while preserving the interpretability of analytic measurements, we calibrate ${\bf{A}}_t^e$ into calibrated complete traces ${\bf{Z}}_t^e$ through a lightweight bounded calibration:
\begin{equation}
{\bf{Z}}_t^e
=
{\rm clip}_{[0,1]}
\left[
\overline{\bf{A}}_t^e
+
\lambda_z
\tanh
\left(
{\cal R}_z^e(\overline{\bf{A}}_t^e)
\right)
\right],
\end{equation}where ${\rm clip}_{[0,1]}(\cdot)$ denotes element-wise clipping,  $\overline{\bf{A}}_t^e$ denotes normalized analytic traces, and ${\cal R}_z^e(\overline{\bf{A}}_t^e)
=
{\cal C}_1
(
{\cal C}_3
(
{\cal C}_3
(
\overline{\bf{A}}_t^e
)))$ represents a lightweight residual calibration network composed of convolution blocks ${\cal C}_i$ with $i\times i$ kernels. The coefficient $\lambda_z$ controls the maximum calibration magnitude and is fixed to 0.1. Since the reliability of individual measurements varies across scenes and spatial locations, HPTO further performs adaptive measurement selection within each observer. Given the calibrated complete traces ${\bf{Z}}_t^e$, global measurement statistics are first extracted:
\begin{equation}
{\bf{q}}_t^e
=
[
{\rm mean}({\bf{Z}}_t^e),
{\rm std}({\bf{Z}}_t^e),
\max({\bf{Z}}_t^e)
].
\end{equation}

The global statistics provide scene-level priors, while the complete traces provide spatial responses. They are jointly processed by a global selection module ${\cal G}_1^e(\cdot)$ and a local selection module ${\cal G}_2^e(\cdot)$ to estimate measurement weights ${\boldsymbol\alpha}_t^e$:
\begin{equation}
{\boldsymbol\alpha}_t^e
=
\frac{
\exp
(
{\cal G}_1^e({\bf{q}}_t^e)
+
{\cal G}_2^e({\bf{Z}}_t^e)
)
}
{
{\rm Sum}_{m}
[
\exp
(
{\cal G}_1^e({\bf{q}}_t^e)
+
{\cal G}_2^e({\bf{Z}}_t^e)
)
]
}.
\end{equation}
Here, ${\cal G}_1^e(\cdot)$ and ${\cal G}_2^e(\cdot)$ are implemented using a lightweight MLP and convolution blocks, respectively, and ${\rm Sum}_{m}(\cdot)$ denotes summation along the $J$ measurement dimensions. The observer summaries ${\bf{F}}_t^e$ is obtained by aggregating selected measurements:
\begin{equation}
{\bf{F}}_t^e
=
{\cal C}_1
(
{\cal C}_3^{\rm dw}
(
{\cal C}_1
(
{\boldsymbol\alpha}_t^e
\odot
{\bf{Z}}_t^e
)
)
),
\end{equation}
where ${\cal C}_3^{\rm dw}$ denotes a convolution block with a $3\times3$ depthwise convolution kernel. The calibrated complete traces ${\bf{Z}}_t^e$ preserve dense spatial responses as fine-grained decision evidence, whereas the summaries ${\bf{F}}_t^e$ provide compact routing representations for subsequent consensus.

\subsection{Evidence-Preserving Dense Consensus}

Although summaries ${\bf{F}}_t^e$ provide compact routing information, compression and nonlinear encoding may weaken responses that remain valuable for reliability estimation. To avoid premature evidence loss, ANCHOR introduces EPDC. It performs observer-level routing by jointly modeling the anchor--proposal relationship and observer responses. The routing weights ${\boldsymbol\rho}_t^e$ among trace families are computed as:
\begin{equation}
{\boldsymbol\rho}_t^e
=
\frac{
\exp
\left(
{\cal S}_e^{\rm rel}({\bf{R}}_t)
+
{\cal S}_e^{\rm obs}({\bf{F}}_t^e)
\right)
}
{
\sum\limits_{e'\in{\cal E}}
\exp
\left(
{\cal S}_{e'}^{\rm rel}({\bf{R}}_t)
+
{\cal S}_{e'}^{\rm obs}({\bf{F}}_t^{e'})
\right)
},
\quad e\in{\cal E},
\end{equation}
where ${\cal S}_e^{\rm rel}(\cdot)$ estimates the relevance of trace family $e$ based on the local anchor--proposal relationship and ${\cal S}_e^{\rm obs}(\cdot)$ evaluates the observer response. Both functions are implemented using lightweight projection layers and convolution blocks.

To construct the evidence representation ${\bf{E}}_t$ while preserving heterogeneous trace details, EPDC retains complete traces as dense decision evidence rather than using compressed summaries. The routed traces are scaled by $|{\cal E}|$ to preserve the original response scale under uniform routing:
\begin{equation}
{\bf{E}}_t
=
{\cal C}_1
\left(
{\cal C}_3
\left(
{\cal C}_3
\left(
[
|{\cal E}|
{\boldsymbol\rho}_t^e
\odot
{\bf{Z}}_t^e
]_{e\in{\cal E}}
\right)
\right)
\right).
\end{equation}
EPDC then integrates the relation feature ${\bf{R}}_t$ and aggregated evidence ${\bf{E}}_t$ through adaptive residual modulation:
\begin{equation}
{\bf{H}}_t
=
{\bf{R}}_t
+
g_c
{\boldsymbol\Gamma}_t
\odot
{\bf{E}}_t ,
\end{equation}
\begin{equation}
{\boldsymbol\Gamma}_t
=
\sigma
\left(
{\cal C}_1
(
{\cal C}_3
(
[
{\bf{R}}_t,
{\bf{E}}_t
]
)
)
\right),
\end{equation}
where $\sigma(\cdot)$ denotes the Sigmoid activation and $g_c$ is a learnable global consensus gain. By decoupling routing signals from dense evidence preservation, EPDC achieves adaptive consensus while retaining complete physical-trace evidence for subsequent Relative Risk Decoding.

\begin{table*}[t]
\centering
\renewcommand{\arraystretch}{1.1}
\begin{tabular*}{\textwidth}{
@{}
l
c
@{\hspace{4pt}}|@{\hspace{5pt}\extracolsep{\fill}}
c c c c c c
@{}
}
\toprule
\multirow{2}{*}{Method}
& \multirow{2}{*}{Venue}
& \multicolumn{2}{c}{RainSynLight}
& \multicolumn{2}{c}{RainSynComplex}
& \multicolumn{2}{c}{NTURain} \\
\cmidrule(lr){3-4}
\cmidrule(lr){5-6}
\cmidrule(lr){7-8}
& 
& PSNR & SSIM
& PSNR & SSIM
& PSNR & SSIM \\
\midrule

ESTINet \cite{zhang2022enhanced}
& TPAMI22
& 36.12
& 0.9631
& 28.48
& 0.8242
& 37.48
& 0.9700 \\

MFGAN \cite{yang2021recurrent}
& TPAMI22
& 36.99
& 0.9760
& 32.70
& 0.9357
& 38.92
& 0.9764 \\

DRSformer \cite{chen2023learning}
& CVPR23
& 36.84
& 0.9739
& 31.61
& 0.9258
& 36.93
& 0.9591 \\

RainMamba \cite{wu2024rainmamba}
& MM24
& 36.74
& 0.9741
& 32.65
& \underline{0.9361}
& 37.87
& 0.9738 \\

\midrule

DeLiVR \cite{sun2026delivr} 
& ICLR26
& 39.96
& 0.9808
& 31.98
& 0.9188
& 39.02
& 0.9759 \\

DeLiVR + ANCHOR-Guide
&
& \underline{39.98}
& 0.9808
& 31.98
& 0.9188
& 39.03
& 0.9760 \\

DeLiVR + ANCHOR-Correct
&
& \textbf{40.08}
& 0.9821
& 31.99
& 0.9193
& 39.04
& 0.9760 \\

\midrule

VDMamba \cite{sun2025semi}
& CVPR25
& 38.67
& \underline{0.9822}
& \underline{33.26}
& \textbf{0.9481}
& 39.81
& \underline{0.9796} \\

VDMamba + ANCHOR-Guide
&
& 38.80
& 0.9821
& \textbf{33.40}
& \textbf{0.9481}
& \underline{39.91}
& \textbf{0.9799} \\

VDMamba + ANCHOR-Correct
&
& 39.20
& \textbf{0.9837}
& \textbf{33.40}
& \textbf{0.9481}
& \textbf{40.07}
& \underline{0.9796} \\

\bottomrule
\end{tabular*}
\caption{Quantitative comparisons on the RainSynLight
\cite{liu2018erase}, RainSynComplex \cite{liu2018erase}, and
NTURain \cite{chen2018robust} datasets. The best and second-best
distinct restoration results are highlighted in bold and underlined,
respectively.}
\label{tab:deraining}
\end{table*}

\begin{table*}[t]
\centering
\renewcommand{\arraystretch}{1.1}

\begin{tabular*}{\textwidth}{
@{}
l
c
@{\hspace{3pt}}|@{\hspace{4pt}\extracolsep{\fill}}
c c c c c c
@{}
}
\toprule
\multirow{2}{*}{Method}
& \multirow{2}{*}{Venue}
& \multicolumn{3}{c}{DeepHDRVideo}
& \multicolumn{3}{c}{Cinematic} \\
\cmidrule(lr){3-5}
\cmidrule(lr){6-8}
&
& PSNR$_T$
& SSIM$_T$
& HDR-VDP-2
& PSNR$_T$
& SSIM$_T$
& HDR-VDP-2 \\
\midrule

Chen \cite{chen2021hdr}
& ICCV21
& 43.32
& \underline{0.9551}
& 78.37
& 39.27
& 0.9168
& 71.64 \\

LAN-HDR \cite{chung2023lan}
& ICCV23
& 41.83
& 0.9499
& 76.00
& 38.22
& 0.9100
& 69.09 \\

LRHDR \cite{liao2026lrhdr}
& CVPR26
& 43.49
& \textbf{0.9630}
& -
& \underline{41.11}
& \textbf{0.9274}
& - \\

F2HDR \cite{Yue_2026_CVPR}
& CVPR26
& \underline{43.87}
& 0.9573
& \textbf{78.93}
& 39.36
& 0.9185
& 71.53 \\

\midrule

HDRFlow \cite{xu2024hdrflow} 
& CVPR24
& 43.25
& 0.9520
& 78.07
& 39.30
& 0.9156
& 71.55 \\

HDRFlow + ANCHOR-Guide
&
& 43.35
& 0.9525
& 78.03
& 39.70
& 0.9174
& 71.41 \\

HDRFlow + ANCHOR-Correct
&
& 43.41
& 0.9523
& 78.29
& 39.66
& 0.9160
& \textbf{71.95} \\

\midrule

NECHDR\cite{cui2024exposure}
& MM24
& 43.44
& 0.9558
& 78.21
& 40.59
& 0.9241
& 70.88 \\

NECHDR + ANCHOR-Guide
&
& \textbf{44.12}
& \underline{0.9599}
& \underline{78.77}
& 41.04
& 0.9238
& 70.96 \\

NECHDR + ANCHOR-Correct
&
& 43.84
& 0.9595
& 78.53
& \textbf{41.31}
& \underline{0.9243}
& \underline{71.74} \\

\bottomrule
\end{tabular*}
\caption{Quantitative comparisons on the DeepHDRVideo \cite{chen2021hdr} and Cinematic \cite{froehlich2014creating} datasets.}
\label{tab:hdr}
\end{table*}

\begin{table}[t]
\centering
\begin{tabular*}{\columnwidth}{
@{\extracolsep{\fill}}
llccc
@{}
}
\toprule
Network & Variant & Params (M) & TFLOPs & FPS \\
\midrule
\multirow{3}{*}{DeLiVR}
& None    & 6.53 & 1.08 & 1.42 \\
& Guide   & 8.55 & 1.35 & 1.40 \\
& Correct & 8.55 & 1.35 & 1.24 \\
\midrule
\multirow{3}{*}{VDMamba}
& None    & 5.17 & 0.43 & 3.51 \\
& Guide   & 7.19 & 0.56 & 3.51 \\
& Correct & 7.19 & 0.56 & 3.13 \\
\midrule
\multicolumn{2}{l}{ANCHOR only}
& 2.02 & 0.14 & 30.54 \\
\bottomrule
\end{tabular*}
\caption{Complexity of ANCHOR variants on NTURain.}
\label{tab:complexity}
\end{table}

\subsection{Relative Risk Decoder}

Conditioned on ${\bf{H}}_t$, the RRD estimates the spatial reliability of the temporal restoration proposal ${\bf{Y}}_t$ with respect to the current-frame anchor ${\bf{X}}_t$. Instead of explicitly predicting restoration errors or requiring risk annotations, RRD learns the relative trust field ${\bf{K}}_t$ between the proposal and anchor through a U-Net style decoder. RRD then converts relative trust field into anchor correction strength through the Calibrated Anchor Correction:
\begin{equation}
{{\bf{M}}_t}
=
{\rm clip}_{[0,1]}
\left(
s{{\bf{K}}_t}
\right),
\end{equation}
where $s$ is a learnable global correction strength shared across samples and spatial locations within each task. Therefore, reliable proposal regions retain the temporal restoration output, while unreliable regions are adaptively corrected toward the current-frame anchor along the residual direction.

\begin{figure}[!ht]
  \begin{center}
    \centerline{\includegraphics[width=\columnwidth]{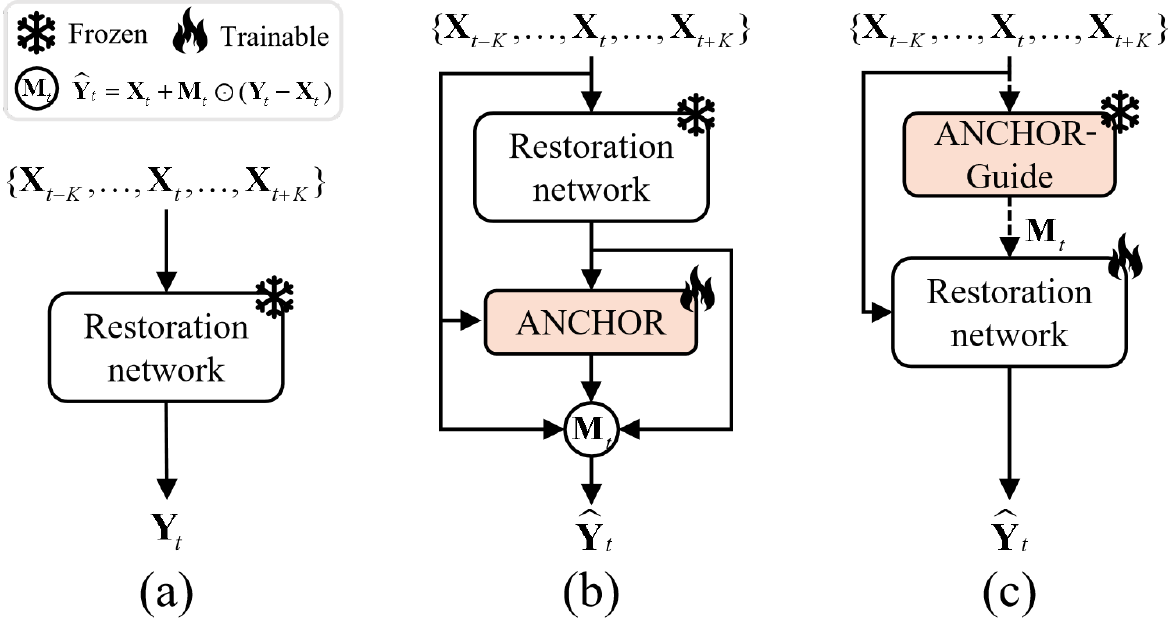}}
    \caption{Pipeline comparison of ANCHOR variants. (a) Restoration network inference. (b) ANCHOR training and inference: ANCHOR-Correct accesses the restoration proposal for output correction, whereas ANCHOR-Guide predicts the correction field without accessing it. (c) ANCHOR-Guide guides restoration network fine-tuning, and the optimized network retains the ANCHOR-Guide configuration in (b).}
    \label{fig:variants}
  \end{center}
\end{figure}

\subsection{Training Objectives}

ANCHOR contains two variants, ANCHOR-Correct and ANCHOR-Guide as illustrated in Fig.~\ref{fig:variants}. They share identical HPTO, EPDC, and RRD architectures but differ in their available inputs. ANCHOR-Correct observes both the low-quality video and the temporal restoration proposal ${{\bf{Y}}_t}$, while ANCHOR-Guide only uses the low-quality video without accessing the restoration proposal.

\textbf{Training ANCHOR.}
We freeze the restoration network and optimize ANCHOR-Correct and ANCHOR-Guide with the objective ${\cal L}$, defined using the original task loss ${\ell}_{\rm Task}$:
\begin{equation}
\begin{aligned}
{\cal L}
&=
{\ell}_{\rm Task}
(
\widehat{\bf{Y}}_t,
{\bf{G}}_t
)
\\
&=
{\ell}_{\rm Task}
(
{\bf{X}}_t
+
{\bf{M}}_t
\odot
(
{\bf{Y}}_t-{\bf{X}}_t
),
{\bf{G}}_t
).
\end{aligned}
\end{equation}

\textbf{Restoration Optimization with ANCHOR-Guide.}
After training, ANCHOR-Guide provides anchor-relative spatial weights through the restoration fine-tuning objective ${\cal L}_{\rm guide}$:
\begin{equation}
{\cal L}_{\rm guide}
=
\frac{
\left\langle
{\bf{M}}_t,
{\ell}_{\rm Task}({\bf{Y}}_t,{\bf{G}}_t)
\right\rangle
}
{
\left\langle
{\bf{M}}_t,
{\bf{1}}
\right\rangle
+
\zeta
}.
\end{equation}
Here, $\langle\cdot,\cdot\rangle$ denotes the tensor inner product, and $\zeta$ ensures numerical stability. ANCHOR-Guide remains frozen during fine-tuning, while ${\bf{M}}_t$ emphasizes regions with unreliable anchors and down-weights well-preserved regions to avoid unnecessary modifications. At inference, the optimized restoration network and ANCHOR-Guide run in parallel, and their outputs are combined to produce the final correction.

\begin{figure*}[!ht]
  \centering
  \includegraphics[width=\textwidth]{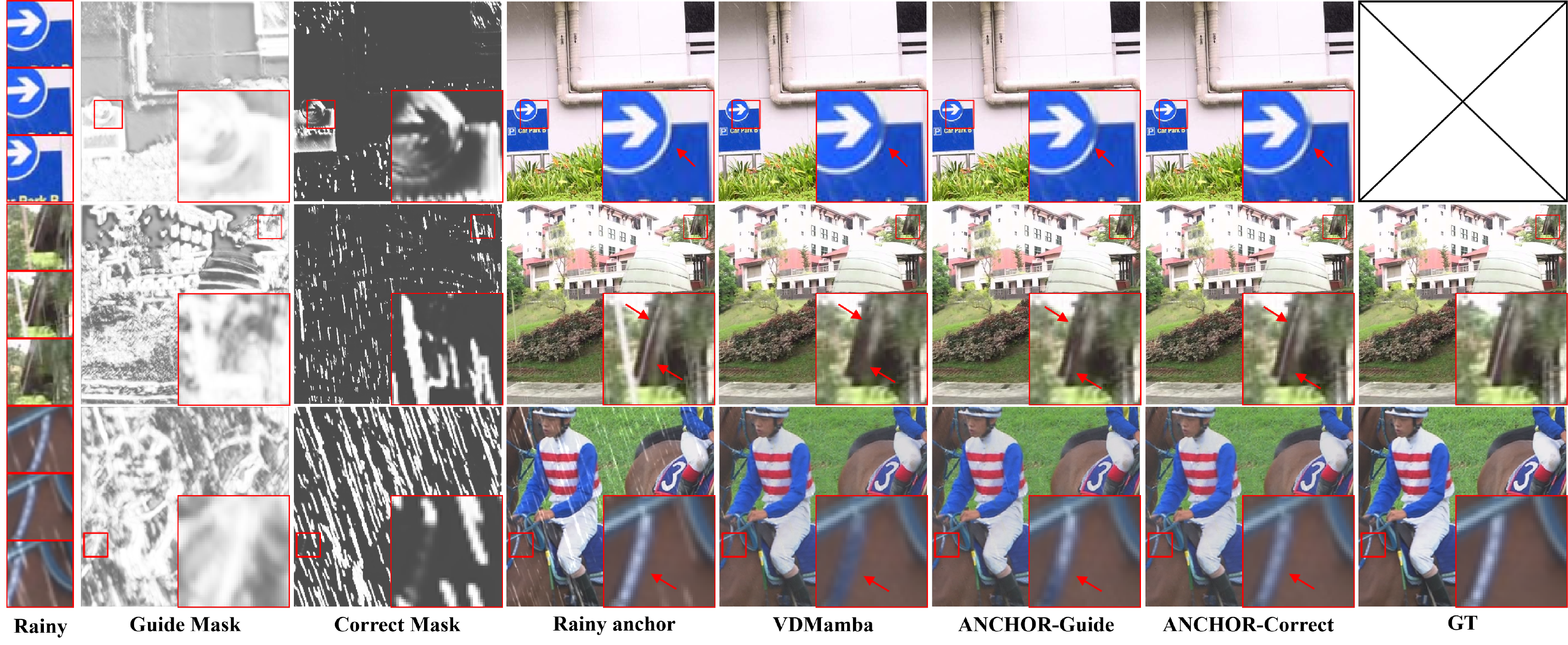}
 \caption{ The qualitative comparisons on datasets RainSynLight
\cite{liu2018erase} and
NTURain \cite{chen2018robust}.}
  \label{fig:deraining}
\end{figure*}
\begin{figure*}[!ht]
  \centering
  \includegraphics[width=\textwidth]{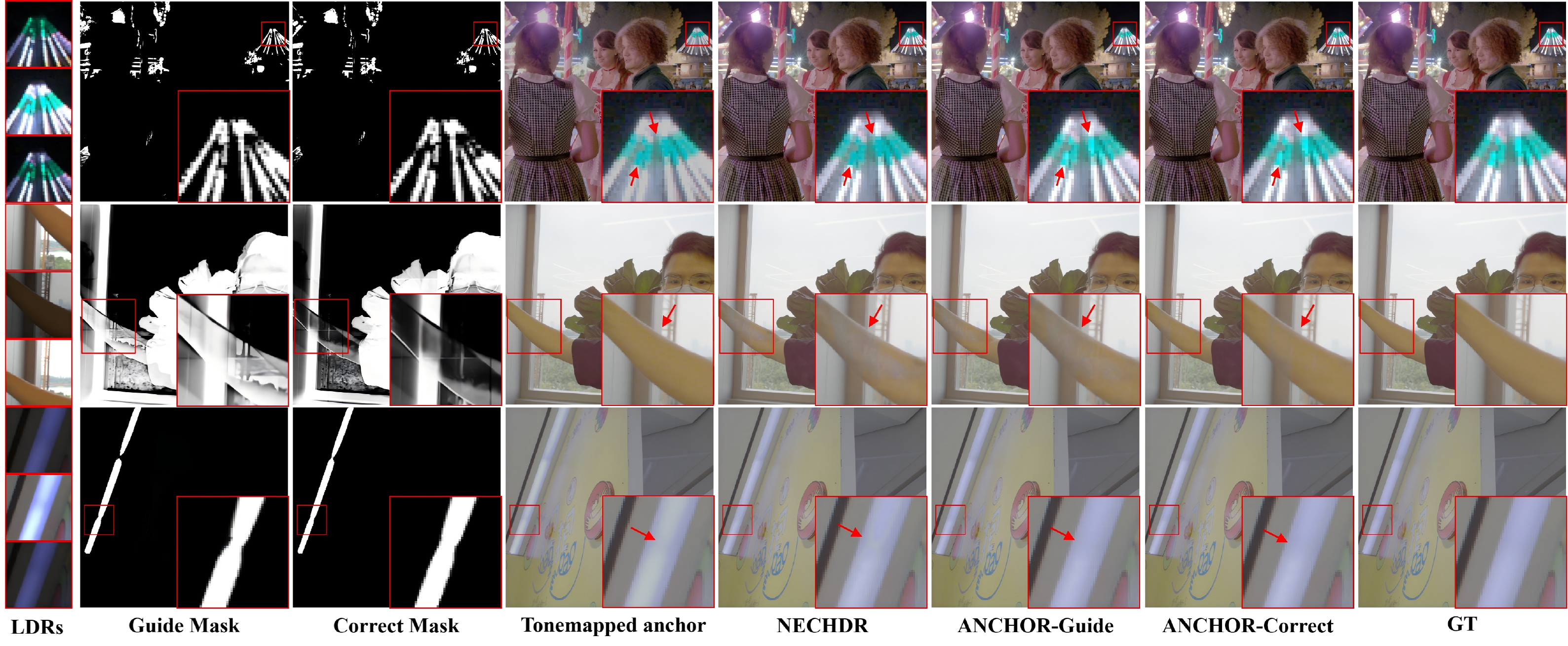}
 \caption{The qualitative comparisons on datasets DeepHDRVideo \cite{chen2021hdr} and Cinematic \cite{froehlich2014creating}.}
  \label{fig:hdr}
\end{figure*}

\section{Experiments}

\subsection{Experimental Settings}

\textbf{Datasets and Baselines.}
We evaluate ANCHOR on video deraining and HDR video reconstruction. For video deraining, we use RainSynLight \cite{liu2018erase}, RainSynComplex \cite{liu2018erase}, and NTURain \cite{chen2018robust}. For HDR video reconstruction, we train on Vimeo-90K \cite{xue2019video} following the protocols of \cite{xu2024hdrflow,cui2024exposure,Yue_2026_CVPR}, and evaluate on DeepHDRVideo \cite{chen2021hdr} and Cinematic \cite{froehlich2014creating}. For video deraining, ANCHOR is integrated into DeLiVR \cite{sun2026delivr} and VDMamba \cite{sun2025semi}, and compared with methods including DRSformer \cite{chen2023learning}, ESTINet \cite{zhang2022enhanced}, MFGAN \cite{yang2021recurrent}, and RainMamba \cite{wu2024rainmamba}. For HDR video reconstruction, ANCHOR is integrated into HDRFlow \cite{xu2024hdrflow} and NECHDR \cite{cui2024exposure}, and compared with F2HDR \cite{Yue_2026_CVPR}, LRHDR \cite{liao2026lrhdr}, LAN-HDR \cite{chung2023lan}, and Chen \cite{chen2021hdr}.

\textbf{Training Details and Metrics.} ANCHOR follows a common training protocol across restoration networks with minor hyperparameter adjustments. For VDMamba, ANCHOR is trained with AdamW for 12 epochs at a learning rate of $5\times10^{-6}$, followed by 10 epochs of Adam at 0.08 to calibrate the global correction strength $s$. For restoration network fine-tuning with ANCHOR-Guide, ANCHOR-Guide is frozen and the network is fine-tuned for 1 epoch at $2\times10^{-8}$. Other networks follow the same protocol with minor adjustments to the learning rate and epochs, detailed settings are provided in the supplementary material. For video deraining, the rainy center frame is directly used as the observation anchor. For HDR video reconstruction, the Low Dynamic Range (LDR) center frame is linearly mapped following prior work \cite{xu2024hdrflow} before serving as the anchor. All experiments are conducted on four NVIDIA RTX 3090 GPUs with 24 GB memory each. We report PSNR and SSIM for video deraining, and PSNR$_T$, SSIM$_T$, and HDR-VDP-2 \cite{mantiuk2011hdr} for HDR video reconstruction. PSNR$_T$ and SSIM$_T$ are computed in the $\mu$-law tone-mapped domain. For fair comparison, all HDR-VDP-2 results use a fixed 28-inch display diagonal and a 55 cm viewing distance. Parameters, FLOPs, and FPS quantify ANCHOR's computational overhead.

\subsection{Quantitative Comparisons}

\textbf{Video Deraining.}
Table~\ref{tab:deraining} reports quantitative comparisons on three video deraining benchmarks. By refining the outputs of DeLiVR and VDMamba, ANCHOR-Correct achieves consistent improvements, with larger gains observed on VDMamba. Meanwhile, ANCHOR-Guide further improves both restoration networks through reliability-guided optimization, demonstrating the effectiveness of training-time correction guidance. Overall, ANCHOR achieves competitive performance against existing state-of-the-art deraining methods.

\textbf{HDR Video Reconstruction.}
Table~\ref{tab:hdr} reports quantitative comparisons on two HDR video reconstruction benchmarks. Both ANCHOR variants improve HDRFlow and NECHDR. In particular, ANCHOR-Guide guides the restoration network to focus optimization on severely degraded regions while avoiding unnecessary modifications to regions with reliable current-frame observations, yielding substantial gains for NECHDR, especially on DeepHDRVideo dataset.

\textbf{Complexity Analysis.}
Table~\ref{tab:complexity} reports the additional complexity of ANCHOR on NTURain. ANCHOR itself runs substantially faster than the restoration networks. Consequently, ANCHOR-Correct causes only a modest FPS reduction, while ANCHOR-Guide runs in parallel without waiting for the restoration proposal and has negligible impact on inference speed across both evaluated restoration networks.

\subsection{Qualitative Comparisons}
Figures~\ref{fig:deraining} and \ref{fig:hdr} compare representative restoration networks with and without ANCHOR. Additional results for DeLiVR and HDRFlow are provided in the supplementary material.

\textbf{Video Deraining.}
As shown in Fig.~\ref{fig:deraining}, ANCHOR mitigates over-smoothing in VDMamba and recovers fine structures lost during temporal restoration. Although temporal aggregation exploits complementary reference-frame information for rain removal, uncertain evidence may suppress reliable high-frequency details along with rain patterns. By evaluating the proposal against the current-frame anchor, ANCHOR restores weakened structures without compromising rain removal, better preserving roof edges and stripe-like patterns. This highlights the value of the degraded current frame as aligned structural evidence for output correction.

\textbf{HDR Video Reconstruction.}
As shown in Fig.~\ref{fig:hdr}, ANCHOR reduces brightness errors, ghosting, and other artifacts induced by NECHDR in dynamic and static scenes with varying exposures. These errors arise from inconsistent reference-frame observations, inaccurate cross-frame correspondences, unreliable feature fusion, and model inductive biases. Rather than globally replacing the restoration prediction, ANCHOR revisits the tone-mapped current-frame anchor to localize correction, integrating direct target-frame evidence while retaining recovered HDR information.

\begin{table}[t]
\centering
\setlength{\tabcolsep}{3pt}
\renewcommand{\arraystretch}{1.05}
\begin{tabularx}{\columnwidth}{
    @{}
    >{\raggedright\arraybackslash}X
    *{4}{>{\centering\arraybackslash}p{0.135\columnwidth}}
    @{}
}
\toprule
\multirow{2}{*}{Method}
& \multicolumn{2}{c}{RainSynLight}
& \multicolumn{2}{c}{DeepHDRVideo} \\
\cmidrule(lr){2-3}
\cmidrule(lr){4-5}
& PSNR & SSIM & PSNR$_T$ & SSIM$_T$ \\
\midrule
VDMamba/NECHDR
& 38.67 & 0.9822 & 43.44 & 0.9558 \\

\midrule

w Concatenation
& 38.63 & 0.9820 & \underline{43.69} & \underline{0.9594} \\
w Addition
& \underline{39.13} & \underline{0.9835} & 43.54 & 0.9572 \\
w EPDC
& \textbf{39.20} & \textbf{0.9837}
& \textbf{43.84} & \textbf{0.9595} \\
\bottomrule
\end{tabularx}
\caption{Ablation study of evidence injection strategies in ANCHOR-Correct on RainSynLight and DeepHDRVideo.}
\label{tab:ablation_deraining2}
\end{table}

\begin{table}[t]
\centering
\setlength{\tabcolsep}{3pt}
\renewcommand{\arraystretch}{1.05}
\begin{tabularx}{\columnwidth}{
    @{}
    >{\raggedright\arraybackslash}X
    *{4}{>{\centering\arraybackslash}p{0.135\columnwidth}}
    @{}
}
\toprule
\multirow{2}{*}{Method}
& \multicolumn{2}{c}{RainSynLight}
& \multicolumn{2}{c}{DeepHDRVideo} \\
\cmidrule(lr){2-3}
\cmidrule(lr){4-5}
& PSNR & SSIM & PSNR$_T$ & SSIM$_T$ \\
\midrule
VDMamba/NECHDR
& 38.67 & 0.9822 & 43.44 & 0.9558 \\

ANCHOR-Correct
& \textbf{39.20} & \textbf{0.9837}
& \textbf{43.84} & \textbf{0.9595} \\
\midrule
w/o HPTO \& EPDC
& 38.67 & 0.9821 & 43.43 & 0.9558 \\

w/o Radiometric 
& \underline{39.14} & 0.9834 & 43.45 & 0.9558 \\

w/o Structural 
& 39.08 & 0.9832 & 43.79 & \underline{0.9593} \\

w/o Temporal 
& 39.13 & \underline{0.9835} & \underline{43.82} & \textbf{0.9595} \\

\bottomrule
\end{tabularx}
\caption{Ablation of HPTO in ANCHOR-Correct on RainSynLight and DeepHDRVideo. Removing HPTO also disables EPDC because no physical-trace evidence remains.}
\label{tab:ablation_deraining}
\end{table}

\begin{figure}[!ht]
  \begin{center}
    \centerline{\includegraphics[width=\columnwidth]{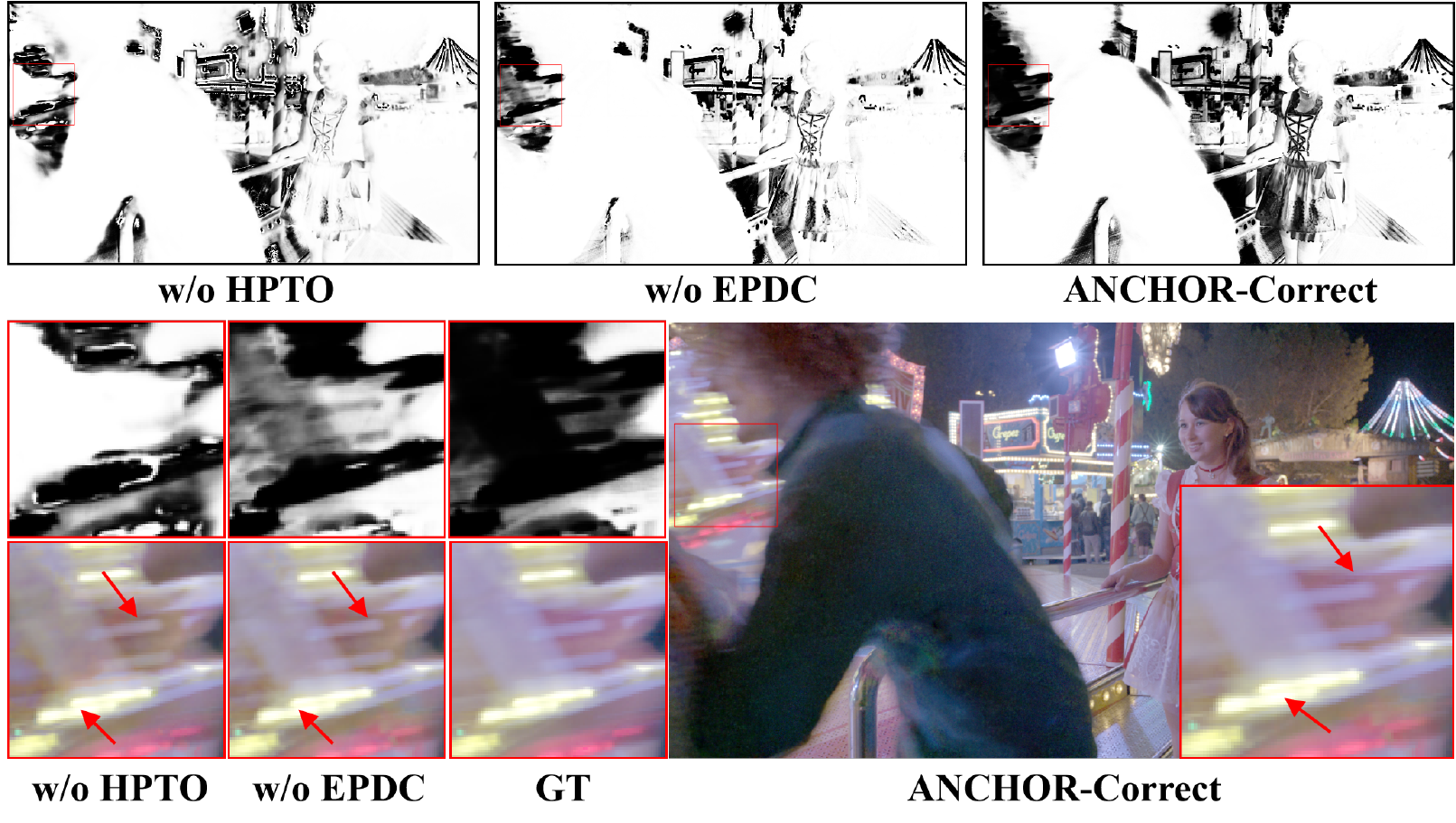}}
    \caption{Qualitative ablation comparison on the Cinematic.
}
\vspace{-1em}
    \label{abhdr}
  \end{center}
\end{figure}

\subsection{Ablation Studies}

We evaluate the contributions of ANCHOR components on RainSynLight and DeepHDRVideo, with further analysis of HPTO and EPDC on the Cinematic dataset, as shown in Tables~\ref{tab:ablation_deraining2} and \ref{tab:ablation_deraining} and Fig.~\ref{abhdr}. HPTO and EPDC play complementary roles in physical-trace extraction and evidence injection, respectively. Replacing EPDC with direct concatenation or addition consistently degrades performance, demonstrating the importance of evidence-preserving aggregation. Removing individual observers also reduces performance, confirming that heterogeneous physical traces provide complementary reliability cues. When HPTO is removed, EPDC is disabled because no physical-trace evidence remains, yielding a relation-only baseline that decodes the correction field solely from anchor--proposal features. This reduced model degenerates to the uncorrected output on RainSynLight and DeepHDRVideo. Since the restoration proposal is generally superior to the degraded anchor, relation features alone provide insufficient evidence to justify correction toward the anchor, whose incorrect use may incur large reconstruction errors. Consequently, the model either conservatively preserves the proposal or produces unreliable corrections. These results demonstrate that HPTO and EPDC are both essential for robust reliability estimation and adaptive output correction.

\section{Conclusion}

We presented ANCHOR, a reliability-aware post-processing framework for video restoration. Without modifying restoration architectures, ANCHOR treats their outputs as restoration proposals and uses the current frame as a temporal anchor to estimate spatial reliability for adaptive correction. The trust-guided correction improves unreliable regions while preserving reliable details, requiring no redesign of the restoration network. Experiments on HDR video reconstruction and video deraining show that ANCHOR consistently enhances diverse state-of-the-art restoration models, demonstrating the effectiveness of reliability-aware post-processing for improving pretrained video restoration systems.




\bigskip

\bibliography{aaai2027}


\end{document}